\pdfoutput=1
\documentclass[11pt]{article}

\usepackage[]{naacl2021}

\usepackage{times}
\usepackage{latexsym}
\usepackage{graphicx}

\usepackage[T1]{fontenc}

\usepackage[utf8]{inputenc}

\usepackage{microtype}

\usepackage{xspace}   
\usepackage{multirow}
\usepackage{booktabs}
\usepackage{array}
\usepackage{arydshln}
\usepackage{graphicx}
\usepackage{soul}   
\usepackage{tabularx}
\usepackage{subcaption}
\usepackage{tabularx}
\usepackage{booktabs}

\usepackage{graphicx}
\usepackage{enumitem}
\usepackage{csvsimple}
\usepackage{dsfont}
\usepackage{pifont}
\usepackage{amsthm}
\usepackage[ruled,vlined]{algorithm2e}
\usepackage{xspace}
\usepackage{amsmath}
\usepackage{arydshln}
\usepackage{multirow}
\usepackage{colortbl}
\usepackage{float}
\usepackage{multicol}
\usepackage{array}
\usepackage{amssymb}
\usepackage{soul}
\usepackage{arydshln}
\usepackage{xcolor}
\newcommand{\hlc}[2][yellow]{{%
    \colorlet{foo}{#1}%
    \sethlcolor{foo}\hl{#2}}%
}

\newcommand\numquestions{5,049\xspace}
\newcommand\numpapers{1,585\xspace}
\newcommand\dataset{\textsc{Qasper}\xspace}
\newcommand\model{LED\xspace}

\title{A Dataset of Information-Seeking Questions \\and Answers Anchored in Research Papers}

\author{Pradeep Dasigi$^\clubsuit$\quad Kyle Lo$^\clubsuit$\quad Iz Beltagy$^\clubsuit$\quad Arman Cohan$^\clubsuit$\quad \\\textbf{Noah A. Smith$^{\diamondsuit\clubsuit}$\quad Matt Gardner$^\clubsuit$}\\
  $^\clubsuit$Allen Institute for AI\quad $^\diamondsuit$Paul G.~Allen School of CSE, University of Washington\\
  \texttt{\{pradeepd,kylel,beltagy,armanc,noah,mattg\}@allenai.org}}

\begin{document}

\maketitle
\begin{abstract}
Readers of academic research papers often read with the goal of answering specific questions. Question Answering systems that can answer those questions can make consumption of the content much more efficient. However, building such tools requires data that reflect the difficulty of the task arising from complex reasoning about claims made in multiple parts of a paper. In contrast, existing information-seeking question answering  datasets usually contain questions about generic factoid-type information. We therefore present \dataset, a dataset of \numquestions{} questions over \numpapers{} Natural Language Processing papers.  Each question is written by an NLP practitioner who read only the title and abstract of the corresponding paper, and the question seeks information present in the full text.  The questions are then answered by a separate set of NLP practitioners who also provide supporting evidence to answers.  We find that existing models that do well on other QA tasks do not perform well on answering these questions, underperforming humans by at least 27 $F_1$ points when answering them from entire papers, motivating further research in document-grounded, information-seeking QA, which our dataset is designed to facilitate.
\end{abstract}

\section{Introduction}
Machines built to assist humans who engage with texts to seek information ought to be designed with an awareness of the information need.  Abstractly, the human's need should define the lens through which the system views the text in order to find desired information. Existing information-seeking machine reading datasets~\cite[e.g.,][]{nq,tydiqa} have led to significant progress in reading at scale~\citep[e.g.,][]{asai2019learning,guu2020realm,liu2020rikinet}. However, most of those benchmarks focus on an ``open domain'' setting where the questions are not anchored in any particular user context.  The result is an emphasis on generic factoid questions, rather than the full range of information needs people have.
\begin{figure}[t]
    \centering
    \includegraphics[width=3in]{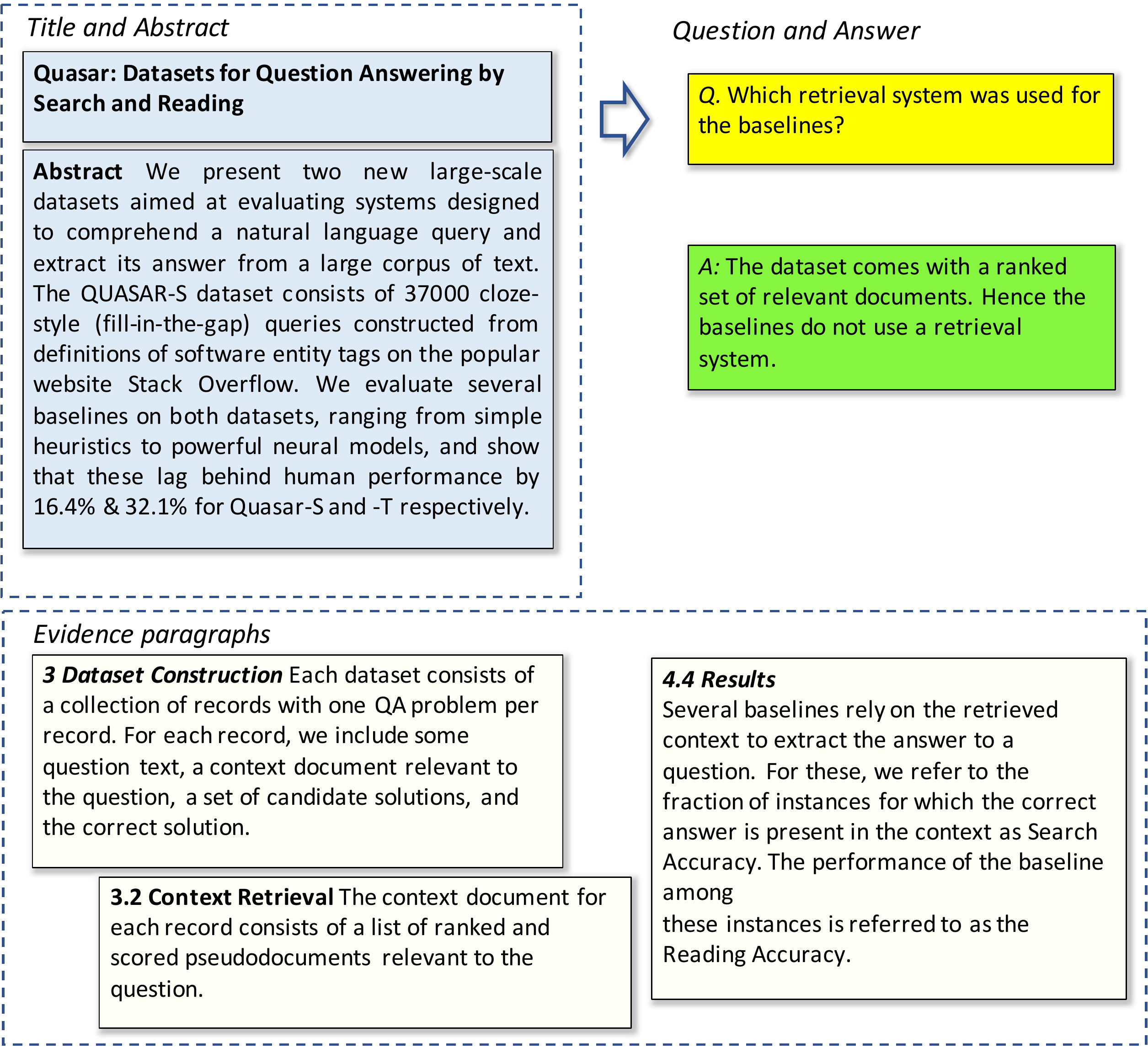}    
    \caption{An example instance taken from \dataset{}. A \hlc[yellow!60]{\textbf{question}} about the paper is written after reading only the title and the abstract. To arrive at the  \hlc[green!40]{\textbf{answer}}, one finds relevant \textbf{evidence}, which can be spread across multiple paragraphs.  In this example, to answer the question about ``baselines'', the reader must realize from evidence from Sections 3 and 4 that ``context documents'' come pre-ranked in the dataset and the paper's ``baselines'' select from these ``context documents.''}
    \label{fig:example}
\end{figure}

We present \dataset{},\footnote{Loosely derived from \textit{Question Answering over Scientific Research Papers}. The dataset, baseline code, and other information about the project can be found at \url{https://allenai.org/project/qasper}.} an information-seeking question answering (QA) dataset over academic research papers.
Each question is written as a follow-up to the title and abstract of a particular paper, and the answer, if present, is identified in the rest of the paper, along with evidence required to arrive at it. This setup results in questions requiring more complex document-level reasoning than prior datasets, because \emph{(i)} abstracts provide rich prompts for questions that can be asked as follow-up and \emph{(ii)}  academic research papers naturally trigger questions by their target readers that require supporting or refuting claims.  This evidence may be spread across the paper, including tables and figures, often resulting in complex entailment problems. The example in Figure~\ref{fig:example} illustrates one such case where we need to retrieve information from paragraphs in three different sections to answer the question.  

\dataset{} contains \numquestions{} questions over \numpapers{} natural language processing (NLP) papers, asked by regular readers of NLP papers, and answered by a separate set of NLP practitioners. Each paper has an average of 3.2 questions, up to a maximum of 12 questions for a single paper. In addition to providing answers when the questions are answerable, the annotators were asked to select text, tables, or figures as evidence required for answering the questions.
55.5\% of the questions require evidence from multiple paragraphs in the paper and 13\% require tables or figures. To the best of our knowledge, \dataset{} is the first QA dataset in the academic research domain focusing on entire papers, and not just abstracts.

To quantify the difficulty of the tasks in \dataset{}, we apply state-of-the-art document-level Transformer \cite{vaswani2017attention} models to the tasks of selecting evidence and generating answers, and show that the best model performance lags behind humans by 27 $F_1$ points at answering questions from entire papers, and 32 $F_1$ points at selecting the paragraphs that provide evidence to answer the questions, indicating that these are both unsolved problems. Additionally, we experiment with oracles that answer questions from gold evidence and find that better pretraining and domain-adaptation might be helpful.

\section{Building the \dataset{} Dataset}
We now describe our process for constructing the dataset. We began with a set of open-access NLP papers, recruited NLP practitioners who are regular readers of research papers, and designed two different data collection interfaces: one for collecting follow-up questions given titles and abstracts, and another for obtaining evidence and answers to those questions.

\subsection{Papers}

We filtered S2ORC \cite{lo-wang-2020-s2orc},\footnote{We accessed both release versions \texttt{20190928} and \texttt{20200705v1}.} a collection of machine-readable full text for open-access papers, to \emph{(i)} those from arXiv with an associated LaTeX source file,\footnote{LaTeX allows us to avoid quality issues with PDF parsing.} and \emph{(ii)}  are in the computational linguistics domain.\footnote{We chose those either tagged with the \texttt{cs.CL} arXiv category or published with an ACL Anthology identifier.} We limited our domain to computational linguistics to ensure high quality as we have access to realistic users through our research network; broader domain collection is left to future work and should be enabled by the proof-of-concept of our protocols given in this paper. 
We used the S2ORC parser (which normalizes multi-file LaTeX sources and resolves comments and macros) to convert LaTeX markup to full text while preserving section and paragraph breaks and math equations. We supplemented the paper text with extracted images of figures and tables associated with their captions; these were crawled from Semantic Scholar.\footnote{\url{http://semanticscholar.org}}
The result of this process was a collection of 18K full text papers for annotation.  

\subsection{Decoupled Data Collection}
To ensure that our questions are realistic, we decoupled the question-writing and question-answering phases. For both tasks we recruited graduate students studying NLP and freelancers practicing NLP through professional networks and Upwork\footnote{\url{https://www.upwork.com/}}.  All the workers were regular readers of NLP papers, and were paid US\$25 per hour on average (\$20-\$40 based on experience). We paid them on a per-hour basis and not a per-question basis to prioritize data quality over quantity. A total of 25 workers wrote questions while 51 answered them.

\paragraph{Questions}
To ensure that annotators were actually interested in the paper they are reading, we provided them with a lightweight search interface to search papers from the aforementioned collection to focus on their papers of interest.
The interface supports entering manual queries and examples of the queries annotators used include
general (e.g., ``computer vision'') or specific (e.g., ``question answering'', ``information extraction'') areas of study, specific tasks (e.g., ``language identification''), entities (e.g., ``bert'', ``transformers'') or concepts (e.g., ``commonsense'', ``interpretability''), or domain 
specifications (e.g., ``medical'', ``wikipedia''). 
Annotators also had the option to not enter any search queries; in this case, they were shown random papers. 
Annotators were displayed only the title and abstracts of relevant papers and asked to write any number of questions they had about the paper.  Annotators were instructed to only write questions that are \emph{not} answerable from the title and abstract but expected to be answered somewhere in the paper.
Annotators also provided basic information about their expertise in NLP and how familiar they already were with the paper for which they asked questions. Most workers (about 70\%) had some experience in NLP, with 20\% having more than five years of experience. A vast majority (94\%) of the abstracts were seen by the question-writers for the first time.  

\paragraph{Answers}

Annotators were randomly assigned papers with all the corresponding questions written for that paper. They were shown the paper title, abstract, question, full text, and all associated figures and tables to answer the questions.  After reading these, annotators were were asked to:

\begin{itemize}
    \item Make a binary decision as to whether the question is answerable given the paper.
    \item If the question is answerable, select the minimal set of \emph{evidence} snippets that contains the answer to the question. This could be (possibly discontiguous) paragraphs from the text and/or figures or tables. Annotators were asked to prioritize text over figures and tables, unless the information required was present only in figures or tables. When multiple paragraphs could serve as evidence, annotators were asked to first prioritize evidence 
    that adequately answered the question, and then paragraphs that occurred earlier in the text.
    \item If the question is answerable, also provide a concise \emph{answer} to the question.  Annotators were also asked to 
    also indicate whether their concise answer was \emph{(i)} extracted from the evidence, \emph{(ii)} ``yes'' or ``no'', or \emph{(iii)} abstractively written.
\end{itemize}
Annotators were allowed to skip
any questions they did not feel comfortable answering. Since the answering task is significantly more complex than the question-writing task, we designed interactive tutorials and qualification exams for the workers for this task using CrowdAQ~\citep{Ning2020EasyRA}. Workers who scored well were invited to work on the task. If the test performance indicated that the workers did not have sufficient NLP knowledge, or were not used to reading papers we did not let them work on the task. In cases where the workers misunderstood the task, but had sufficient background knowledge, we provided additional training before letting them work on the task.  

\section{\dataset{} Analysis}\label{sec:analysis}
Table~\ref{tab:qasper_examples_answer_evidence} provides representative examples from \dataset{} categorized by question, answer, and evidence types, which we describe here in greater detail.

\begin{table*}[h!]
\footnotesize
\centering

\begin{tabular}{@{}p{1.8in}p{2.2in}p{.5in}p{.3in}p{.6in}@{}}\\
\toprule
\multicolumn{2}{l}{\textbf{Question}} & \textbf{Type} & \textbf{\%}  & \textbf{Paper(s)} \\
\midrule
 \multicolumn{2}{l}{What datasets do they use?}  & General & 33.3\% & \href{https://arxiv.org/abs/1707.06519}{1}; \href{https://arxiv.org/abs/1811.08603}{2}; \href{https://arxiv.org/abs/1906.08286}{3}  \\
 \hdashline[0.4pt/2pt]\noalign{\vskip 0.5ex} 
 \multicolumn{2}{l}{What other political events are included in the database?}  & Specific & 66.7\% & \href{https://arxiv.org/abs/1706.01875}{1706.01875} \\
\bottomrule
\toprule
\textbf{Question} &  \textbf{Answer} & \textbf{Type} & \textbf{\%}  & \textbf{Paper} \\
\midrule

 What five dialogue attributes were analyzed? & Model; Confidence; Continuity; Query-relatedness; Repetitiveness; Specificity & Extractive & 51.8\% & \href{https://arxiv.org/abs/1705.00571}{1705.00571} \\
 \hdashline[0.4pt/2pt]\noalign{\vskip 0.5ex} 
 Which neural architecture do they use as a base for their attention conflict mechanisms? & GRU-based encoder, interaction block, and classifier consisting of stacked fully-connected layers. & Abstractive & 24.2\% & \href{https://arxiv.org/abs/1906.08593}{1906.08593} \\
 \hdashline[0.4pt/2pt]\noalign{\vskip 0.5ex} 
 Do they ensure the that the architecture is differentiable everywhere after adding the Hungarian layer? & Yes & Yes/No & 13.9\% & \href{https://arxiv.org/abs/1712.02555}{1712.02555} \\
 \hdashline[0.4pt/2pt]\noalign{\vskip 0.5ex} 
What language are the captions in? & N/A & Unanswer. & 10.2\% & \href{https://arxiv.org/abs/1909.09070}{1909.09070} \\
\bottomrule 
\toprule
 \textbf{Question} &  \textbf{Evidence} & \textbf{Type} &\textbf{\%} & \textbf{Paper} \\
\midrule
 What new tasks do they use to show the transferring ability of the shared meta-knowledge?
 & To test the transferability of our learned Meta-LSTM, we also design an experiment, in which we take turns choosing 15 tasks to train our model with multi-task learning, then the learned Meta-LSTM are transferred to the remaining one task. The parameters of transferred Meta-LSTM, $\theta^{(s)}_m$ in Eq.( 33 ), are fixed and cannot be updated on the new task. & Text & 81.6\% & \href{https://arxiv.org/abs/1802.08969}{1802.08969} \\
 \hdashline[0.4pt/2pt]\noalign{\vskip 0.5ex} 
 How much does it minimally cost to fine-tune some model according to benchmarking framework? & \href{https://www.semanticscholar.org/paper/HULK\%3A-An-Energy-Efficiency-Benchmark-Platform-for-Zhou-Chen/c26f90d4cfa33ceff373cf49c2a534e2004685da/figure/0}{Table 1} & Table/Figure & 11.6\% & \href{https://arxiv.org/abs/2002.05829}{2002.05829} \\
 \hdashline[0.4pt/2pt]\noalign{\vskip 0.5ex} 
 Do they recommend translating the premise and hypothesis together? & N/A & None & 12.8\% & \href{https://arxiv.org/abs/2004.04721}{2004.04721} \\
\bottomrule
\end{tabular}
\caption{Examples of questions (top), answers (middle), and evidence (bottom) sampled from \dataset. \% are relative frequencies of the corresponding type over all examples in \dataset. The percentages for evidence types sum over 100\% due to double-counting of 446 answers with both Table/Figure and Text evidence.}
\label{tab:qasper_examples_answer_evidence}
\end{table*}

\paragraph{Question types}
We first analyze whether our annotation setup results in questions that are anchored in the context of the papers. To answer this question, we manually\footnote{Two domain-experts independently judged these, and achieved a Cohen's $\kappa$ of 0.94.} categorized a set of 200 questions as being applicable to most papers in the domain (general) vs.~being applicable only to the paper that the question is written about (specific).
Table~\ref{tab:qasper_examples_answer_evidence} shows that most of the questions (67\%) are specific to the papers they are written about. This result indicates the advantage of viewing the \dataset{} task as a question answering problem, instead of an information extraction problem since a fixed schema would not be able to handle the long tail of paper-specific information needs.

\paragraph{Answer types}
As shown in Table~\ref{tab:qasper_examples_answer_evidence}, most of the answers in the dataset are extractive. The average length of the extractive answers is 14.4 words (including all spans), and that of abstractive spans is 15.6 words.

\paragraph{Evidence types}
Evidence can include one or more paragraphs from the paper, a figure, or a table, or a combination of these. Table~\ref{tab:qasper_examples_answer_evidence} shows the distribution of these types. Among the answerable questions with text-only evidence, 55.5\% of the answers have multi-paragraph evidence (Figure~\ref{fig:example} is one example). Unanswerable questions do not have any evidence. Among the answerable ones, (3.0\%) have no evidence when the answer is \textit{No}, and the evidence is the \emph{lack} of a mention of something specific. The last question in Table~\ref{tab:answer_from_evidence_results} is one example of such a case.

\paragraph{Distribution of evidence paragraphs}  We perform an analysis to identify the main sections of a paper that contain textual evidence. We assign each evidence paragraph to its containing top-level\footnote{S2ORC provides section hierarchy derived from LaTeX source} section, and perform some section name normalization. We find that among the frequently used section names such as ``Experiments'' and ``Introduction,'' there was not a single section name that contained a majority of evidence spans, indicating that the distribution of evidence over section in the paper was more or less uniform.

\paragraph{Inter-annotator agreement} 
44\% of the questions in \dataset have multiple annotated answers. 
On average, each question is answered by 1.6 annotators (up to a maximum of 6 annotators for the same question). Using these multiple annotations, we compute some measures of agreement between annotators. First, we found that there is a high level of agreement (90\%) regarding answerability of questions. Second, we find that annotators agreed on the type of the evidence (text vs.~figure) in 84.0\% of the cases. Papers often provide the same information both in tables and text, and agreement over the evidence types could be a consequence of our clear annotation guidelines regarding selecting evidence.

\paragraph{Correctness} To estimate the correctness of the answer annotations in \dataset{}, we manually analyzed 100 randomly sampled questions with multiple answer annotations (averaging 2.73 answers per question). We found that 207 (75.8\%) of the answers were correct. 98\% of the questions had at least one correct answer, and 77\% had most of the answers correct.

\section{Modeling \dataset{}}
This section explains the task, evaluation metrics, and a model addressing \dataset tasks.

\subsection{Task Setup}
We formally define the \dataset{} tasks as follows: Given a paper, and a question about it, the primary task is to determine if the question is answerable, and output a predicted answer, that is one or more spans in the full-text of the paper, \textit{yes}, \textit{no} or other free-form text. A system built for this will be evaluated based on the correctness of the predicted answer measured against the reference answers. Since \dataset{} also provides labeled evidence for all questions, the system may also use auxiliary supervision provided by the evidence.

One such auxiliary task is to predict the evidence required for the question. The inputs are the same as that of the primary task, but the outputs are expected to be one or more paragraphs in the full-text, figures, or tables, and they will be evaluated against labeled evidence spans.

\paragraph{Evaluation metrics}
As an automatic proxy for the measure of correctness of all types of answers, we use the span-level $F_1$ measure proposed by~\citet{Rajpurkar2016squad}. We convert answers that are multiple selected spans into single comma-separated strings. For questions with multiple reference answers, we compute the max span-$F_1$ of the predictions over all the references. We evaluate the performance of a system over the auxiliary task by computing a $F_1$ score over the set of paragraphs, figures, and tables chosen by the system against the reference evidence, considering a max when there are multiple references. We refer to these metrics as Answer-$F_1$ and Evidence-$F_1$, respectively.

\paragraph{Data splits}
We split the dataset into train, validation, and test sets, so that each paper appears in only one of them. Our analysis of correctness of annotations presented in Section~\ref{sec:analysis} indicates a high likelihood (98\%) of evaluating against a correct reference when evaluation is aggregated over multiple references. Hence we ensure that most of the questions in validation and test sets have multiple references (98\% in test, and 74\% in validation). This resulted in 2,593, 1,005, and 1,451 questions in the three sets, respectively.

\paragraph{Estimating human performance}
To estimate an upper bound on model performance given our data splits and metrics, we assess the performance of the workers when evaluated against each other using the same metrics on a sample of the test set. Since model performance is evaluated by aggregating over multiple references, we consider a subset of the test set containing questions with at least three references (40\% of the test set), evaluate each reference against the remaining, and compute an average over all such combinations. This procedure estimates the human performance to be 60.9 Answer-$F_1$, and 71.6 Evidence-$F_1$. Note that given the disagreements among the workers estimated in Section~\ref{sec:analysis}, this is a lower bound on human performance for two reasons: first, because only two annotations are used to compute the metric, while systems are evaluated against all three; and second, because the annotators are NLP practitioners, not expert researchers, and it is likely that an expert would score higher.
Hence we report these numbers, along with a breakdown over answer types in Table~\ref{tab:answer_results} and Table~\ref{tab:evidence_results} as human performance lower bounds.

\subsection{\dataset Model}

We base our model on pretrained Transformer \cite{vaswani2017attention} models which currently produce state-of-the-art results on a majority of QA tasks.\footnote{\url{https://paperswithcode.com/task/question-answering}} 
Recall that \dataset introduces two main modeling challenges -- 
different answer types and long input documents. 

First, \dataset includes a variety of answer types, including extractive, abstractive, yes/no, and unanswerable questions, which means a typical span-selection BERT-based QA model~\cite{bert} is not sufficient to support all these answer types. 
We address this by converting all answer types into a single task: generating answer text~\cite{t5,unifiedqa}.\footnote{We tried a model that predicts answer type, then based on the type uses a different head to predict the corresponding answer. This model performed much worse than the proposed seq2seq formulation.} This is a sequence-to-sequence formulation that requires an encoder-decoder Transformer model where the encoder reads the question and the document and the decoder generates the answer text.

Second, research papers are much longer than the typical 512 or 1024 token limit of most BERT-like models, so 
we need a Transformer model that can process long inputs. 
We use the Longformer-Encoder-Decoder (\model; \citealp{beltagy2020longformer}), an encoder-decoder Transformer model that can efficiently process input sequences thousands of tokens long.
With \model's  support for input sequence length of 16K tokens, we can encode 99\% of the paper full texts in the \dataset dataset without truncation. 

\paragraph{Longformer-Encoder-Decoder (\model)}
\model \cite{beltagy2020longformer} is a variant of the original Transformer encoder-decoder model that replaces the Transformer's full self-attention in the encoder with the efficient local+global attention pattern of Longformer. This allows each token to attend to only its local window and a pre-specified set of global locations of interest, thereby scaling self-attention computation linearly with the input size (as opposed to quadratically with full context self-attention). \model has a similar architecture to BART~\cite{bart} in terms of number of layers and hidden state sizes, with the distinction that it has a larger position embeddings matrix, allowing it to process inputs of up to 16K tokens long (up from 1K tokens in the original BART model). In practice, \model's parameters are initialized from a pretrained BART model, and \model copies BART's position embeddings 16 times to fill the entire 16K position embeddings matrix. For all experiments we use the \model-base sized model, which uses BART-base weights. 

\paragraph{Input and Output Encoding }

For the input, we follow
the Longformer QA models~\cite{beltagy2020longformer} and encode the question and context in one concatenated string with ``global attention'' over all the question tokens.
For the output, all answer types 
are encoded as single strings. 
The string is the text of the abstractive answer, a comma separated concatenation of the extractive spans, ``Yes'', ``No'',  or ``Unanswerable''.

\paragraph{Evidence extraction}
To support extracting evidence paragraphs, we prepend each paragraph with a \texttt{</s>} token and add a classification head over these tokens on \model's encoder side. We also add Longformer's global attention over these tokens to facilitate direct information flow across the paragraphs. 
We then train \model using both loss functions (teacher-forced text generation and paragraph classification) in a multi-task training setup.
For the answer generation, we use a cross-entropy loss function over the vocabulary. For the evidence paragraph extraction, we use a cross-entropy loss function with binary 0 or 1 gold labels for evidence/non-evidence paragraph. To account for class imbalance, we use loss scaling with weights proportional to the ratio of positive to negative gold paragraphs in the batch, which we found to be crucial for the model to train. One benefit of multi-task training of evidence extraction along with answer selection is that tasks can benefit each other (see Section~\ref{subsec:results}).

\section{Experiments}
We evaluate model performance on question answering and evidence selection tasks, and compare them to estimated lower bounds on human performance.  These human performance estimates are calculated by comparing the answers of questions for which we have multiple human annotations. For each question, we choose one annotation as if it were a prediction, and evaluate it against the rest of the annotations, and consider as human performance the average over all annotations chosen as predictions. We restrict our experiments to the subset of questions in \dataset{} that can be answered from text in the paper, ignoring those that require figures or tables as evidence (13\% of the dataset; see Section~\ref{sec:analysis}) to avoid having to deal with multimodal inputs. We leave multimodal question answering to future work.

\begin{table*}[t]
\small
\centering
\renewcommand{\arraystretch}{1.2}
\begin{tabular}{lcccccccccc}
\toprule
\multirow{2}{*}{Input} & \multicolumn{2}{c}{Extractive} & \multicolumn{2}{c}{Abstractive} & \multicolumn{2}{c}{Yes/No} & \multicolumn{2}{c}{Unanswerable} & \multicolumn{2}{c}{Overall} \\
& Dev. & Test & Dev. & Test & Dev. & Test & Dev. & Test & Dev. & Test \\
\midrule
Q only                            & 4.60 & 5.91 & 6.06 & 7.38 & 69.05 & 66.36 & 58.43 & 66.67 & 17.81 & 22.48 \\
Q+Abstract                        & 6.69 & 7.97 & 7.50 & 8.25 & 69.05 & 63.43 & 51.14 & 62.50 & 18.60 & 22.30 \\
Q+Introduction                    & 4.40 & 6.60 & 2.52 & 3.16 & 65.87 & 67.28 & 71.00 & 78.07 & 18.30 & 24.08 \\
\hdashline[.4pt/1pt]
Q+Full Text                      & 26.07 & 30.96 & 16.59 & 15.76 & 67.48 & 70.33 & 28.57 & 26.21 & 29.05 & 32.80 \\
Q+Full Text w/ scaff.      & 24.62 & 29.97 & 13.86 & 15.02 & 63.64 & 68.90 & 38.89 & 44.97 & 28.01 & 33.63 \\
\hdashline[.4pt/1pt]
Human (lower bound) & - & 58.92 & - & 39.71 & - & 78.98 & - & 69.44 & - & 60.92 \\
\bottomrule
\end{tabular}
\caption{\model-base and lower-bound human performance on answering questions in \dataset{}, measured in Answer-$F_!$. The top three rows are heuristic baselines that try to predict answers without encoding entire papers. \textit{w/ scaff.} refers to the inclusion of the evidence selection scaffold during training.}
\label{tab:answer_results}
\end{table*}

\subsection{Training Details}

We train all models using the Adam optimizer~\citep{kingma2014adam} and a triangular learning rate scheduler \cite{howard2018universal} with 10\% warmup. To determine number of epochs, peak learning rate, and batch size, we performed manual hyperparameter search on a subset of the training data. We searched over \{1, 3, 5\} epochs with learning rates \{$1e^{-5}$, $3e^{-5}$, $5e^{-5}$, $9e^{-5}$\}, and found that smaller batch sizes generally work better than larger ones. Our final configuration was 10 epochs, peak learning rate of $5e^{-5}$, and batch size of 2, which we used for all reported experimental settings.  When handling full text, we use gradient checkpointing~\cite{gradckpt} to reduce memory consumption. We run our experiments on a single RTX 8000 GPU, and each experiment takes 30--60 minutes per epoch. 

\subsection{Results}
\label{subsec:results}
\paragraph{Question answering}
Table \ref{tab:answer_results} shows the overall performance of the \model-base model\footnote{We trained an \model-large model as well, but it performed much worse than the base model on the QA task.} on question answering, as well as the performance breakdown on the different answer types.
The table also compares \model-base variants when the input is heuristically limited to smaller parts of the paper (i.e., no context, abstract, introduction).
We generally observe that, by using more context, the performance improves. Specifically, as we observe in row 5 encoding the entire context results in significant overall performance improvement ($\Delta=+9.5$) over the best heuristic (``introduction''). This signifies the importance of encoding the entire paper. 
Comparing rows 4 and 5, we observe that using the evidence prediction as a multi-task scaffolding objective helps, improving the results by $\Delta=+0.8$ points.

\paragraph{Evidence selection}
Table \ref{tab:evidence_results} illustrates the evidence selection
performance of the \model-large and \model-base models compared with simpler baselines. We observe that LED variants outperform the simple TF-IDF baseline but there still remains a large gap to human performance. 

\begin{figure}
    \centering
    \includegraphics[width=3in]{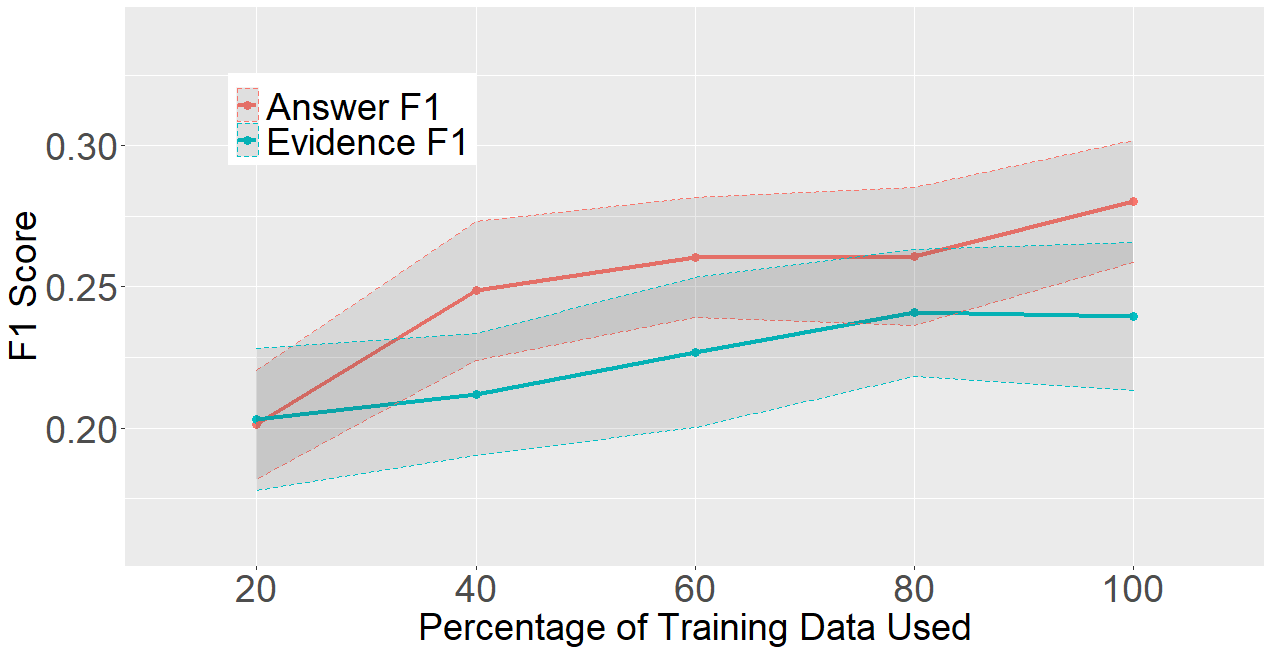}
    \caption{Learning curves showing Answer-$F_1$ and Evidence-$F_1$ on the dev. set while varying training data size.}
    \label{fig:learning_curve}
\end{figure}

\paragraph{Varying amounts of training}
Figure \ref{fig:learning_curve} shows the learning curve that measures the validation Answer-$F_1$ and Evidence-$F_1$ of the \model-base variants based on training data size. The learning curve suggests that performance has not reached a plateau, and future data collection could be useful. 

\paragraph{Answer prediction from gold evidence}
To better isolate the question answering (as opposed to evidence selection) task performance, we perform oracle experiments where models are given the gold evidence.  For these experiments, we are able to use larger \citep[T5-large;][]{t5} or better task-adapted pretrained models \citep[UnifiedQA-large;][]{unifiedqa}, which perform significantly better in the oracle setting. We did not use them in the non-oracle setting, however, as Longformer versions of these models are not available, and LED's ability to handle the full document without the need for a pipelined retrieval system was more important.  These experiments show that (1) the human lower bound is in fact a lower bound, as large models exceed it for span answers in this setting; (2) the majority of the large headroom in the non-oracle setting can be closed with better evidence selection; and (3) research into making large pretrained models able to better scale to long documents would be beneficial.

\begin{table}[t]
\small
\centering
\renewcommand{\arraystretch}{1.2}
\begin{tabular}{lcc}
\toprule
\multirow{2}{*}{Model} & \multicolumn{2}{c}{Evidence $F_1$}\\
& Dev. & Test \\
\midrule
LED-base & 23.94 & 29.85\\
LED-large & 31.25 & 39.37 \\
\hdashline[.4pt/1pt]
TF-IDF & 10.68 & 9.20 \\
Random paragraph & 2.09 & 1.30 \\
First paragraph & 0.71 & 0.34 \\
\hdashline[.4pt/1pt]
Human (lower bound) & - & 71.62  \\
\bottomrule
\end{tabular}
\caption{Model and lower-bound human performance on selecting evidence for questions in \dataset{}}
\label{tab:evidence_results}
\end{table}

\begin{table}[t]
\small
\centering
\renewcommand{\arraystretch}{1.2}
\begin{tabular}{lccccc}
\toprule
\multirow{2}{*}{Model} & \multicolumn{3}{c}{Answer $F_1$} \\
& Span & Abstractive &  Overall \\
\midrule
LED-base  & 54.20 & 24.95  & 44.96 \\
T5-large                    & 65.59 & 29.11 & 60.03  \\
UnifiedQA-large             & 67.23 & 28.92 & 61.39 \\
\bottomrule
\end{tabular}
\caption{Model performance on the \dataset{} test set on answering questions given gold evidence. We do not show performance on \emph{Yes/No} and \emph{Unanswerable} types because they can be trivially predicted to a large extent from the absence of gold evidence.}
\label{tab:answer_from_evidence_results}
\end{table}

\paragraph{Error analysis} To gain insight into the model's errors, we sample 67 test questions with predicted Answer-$F_1$ scores below 0.10 from the \model model trained with evidence prediction scaffolding.  We remove four cases in which the predicted answers are actually correct.  Examining gold answers of the remaining 63, we find 31 are extractive, 24 are abstractive, 3 are ``yes'', 3 are ``no,'' and 2 are unanswerable. We observe that \model often predicts shorter spans than the gold answers (9.5 words shorter than gold counterparts, on average). Focusing only on the 55 questions with either extractive or abstractive gold answers, we manually categorize error types in Table~\ref{tab:error-analysis}.

\begin{table*}[t!]
\centering
\small
\begin{tabular}{@{}lcllll@{}}
\toprule
\textbf{Error} & \textbf{\%} & \textbf{Example question} & \textbf{Gold} & \textbf{Predicted} &  \\ \midrule

\begin{tabular}{@{}l@{}} Incorrectly \\ predicts \\ unanswer- \\able \end{tabular}     &   34.5\%   & \begin{tabular}{@{}l@{}}How is the text segmented? \end{tabular} & \begin{tabular}{@{}l@{}}``dividing documents into chunks \\ before processing'' \end{tabular}
  & Unanswerable    \\

\hdashline[0.4pt/2pt]\noalign{\vskip 0.5ex} 

\multirow{3}{*}{\begin{tabular}{@{}l@{}}Lacks \\ domain \\ knowledge \end{tabular}} 
&  \begin{tabular}{@{}l@{}} \\ 23.6\% \\  \end{tabular}   

&  \begin{tabular}{@{}l@{}}What other scenarios can the bias \\ mitigation methods be applied to?  \end{tabular} & \begin{tabular}{@{}l@{}}``sentiment analysis, other identity \\ problems like racial'' \end{tabular} & GRU  \\ 

& 

& \begin{tabular}{@{}l@{}}What baselines is the neural \\ relation  extractor compared to? \end{tabular}  &  \begin{tabular}{@{}l@{}}Berant et al. (2013), Yao and \\ Van Durme (2014),  Xu et al.\\ (2014), Berant and Liang (2014), ... \end{tabular} & \begin{tabular}{@{}l@{}}Freebase, MCCNN \end{tabular} \\
&
& \begin{tabular}{@{}l@{}} $^\dagger$What hand-crafted features \\ are used? \end{tabular}  &  \begin{tabular}{@{}l@{}}``position of sentence, sentence\\ length, tense, qualifying adjectives,\\ meta-discourse features'' \end{tabular} & \begin{tabular}{@{}l@{}}the Minimum word \\ count is 40, The \\ number of threads \\ to run in parallel is 4 \\ and the context \\ window is 10 \end{tabular} \\
  
\hdashline[0.4pt/2pt]\noalign{\vskip 0.5ex}

\multirow{2}{*}{\begin{tabular}{@{}l@{}}Uninfor- \\ mative  \end{tabular}} 
& \begin{tabular}{@{}l@{}}  \\ 20.0\% \\  \end{tabular}

& \begin{tabular}{@{}l@{}}What do they mean by intrinsic \\ geometry of spaces of learned \\ representations?\end{tabular}  & \begin{tabular}{@{}l@{}} ``the inferred embedding space\\ creates  a globally consistent\\ structured prediction of the\\ ontology, rather than local relation\\ predictions'' \end{tabular} & \begin{tabular}{@{}l@{}}intrinsic  geometry \end{tabular} \\
&
& \begin{tabular}{@{}l@{}}How does the proposed training \\ framework mitigate the bias \\ pattern? \end{tabular}  & \begin{tabular}{@{}l@{}} by balancing or, smoothing the\\ artifacts across different classes by\\ assigning specific weights for every\\ sample \end{tabular} & \begin{tabular}{@{}l@{}} By minimizing the \\ impact of the bias \\ pattern on the dataset \end{tabular} \\

\hdashline[0.4pt/2pt]\noalign{\vskip 0.5ex}

\multirow{2}{*}{\begin{tabular}{@{}l@{}}Not com- \\ prehensive \end{tabular}} 
& \begin{tabular}{@{}l@{}}  \\ 7.3\% \\  \end{tabular} 
 
& \begin{tabular}{@{}l@{}}Which metrics were considered? \end{tabular} & \begin{tabular}{@{}l@{}}``ter, bleu, rouge, nist, lepor, cider,\\ meteor, Semantic Similarity (sim),\\ readability and grammaticality''\\ \end{tabular} & \begin{tabular}{@{}l@{}}Grammar-based \\ metrics (GBMs) \end{tabular} \\
&
& \begin{tabular}{@{}l@{}}  Was permission sought from the \\ bipolar patients to use this data? \end{tabular} & \begin{tabular}{@{}l@{}} For Twitter and Reddit users,\\ implicit consent is assumed to use\\ their public tweets.  Blog users are\\  contacted to obtain consent for\\ using their texts. \end{tabular} & \begin{tabular}{@{}l@{}} No \end{tabular} \\

\hdashline[0.4pt/2pt]\noalign{\vskip 0.5ex}

\multirow{2}{*}{\begin{tabular}{@{}l@{}}Lacks \\ specificity \end{tabular}} 
& \begin{tabular}{@{}l@{}}  \\ 7.3\% \\  \end{tabular} 
& What are the performance metrics? & \begin{tabular}{@{}l@{}}``Rouge-1, Rouge-2 and\\ Rouge-4 recall'' \end{tabular} & \begin{tabular}{@{}l@{}}Rouge scores \end{tabular} \\
& 
& \begin{tabular}{@{}l@{}}What supervised machine \\ learning models do they use? \end{tabular} & \begin{tabular}{@{}l@{}} ``ZeroR, Naïve Bayes, J48, and \\ random forest'' \end{tabular} & \begin{tabular}{@{}l@{}}Weka classifiers \end{tabular} \\

\hdashline[0.4pt/2pt]\noalign{\vskip 0.5ex} 

\begin{tabular}{@{}l@{}} Lacks \\ numeracy \end{tabular}     &   7.3\%   & \begin{tabular}{@{}l@{}}How many tags are included \\ in the ENE tag set? \end{tabular} & ``200 fine-grained categories''   
  & 1     \\

   \bottomrule
\end{tabular}
\caption{Error analysis of our best model (\model from row 5 from Table~\ref{tab:answer_results}) on 55 test examples with low $F_1$ score (excluding those with ``yes,'' ``no,'' or ``unanswerable'' gold answers). ``Quotations'' denote extractive gold answers. We note \emph{Lacks domain knowledge} errors are not always solved by better entity type resolution (see $\dagger$). }\label{tab:error-analysis}
\end{table*}

\section{Related Work}
\paragraph{Information-Verifying QA}
A large body of work on question answering follows the \textit{information-verifying} paradigm where the writer of the question already knows its answer, and the questions are written solely for evaluating the knowledge or understanding capabilities of machines. Some examples include SQuAD  \cite{Rajpurkar2016squad}, TriviaQA \cite{triviaqa}, NarrativeQA \cite{narrativeqa}, WikiHop \cite{wikihop-medihop}, HotpotQA \cite{Yang2018HotpotQA}, CoQA \cite{coqa}, DROP \cite{drop}, \textsc{Quoref} \cite{dasigi-etal-2019-quoref}. Most datasets for QA on academic research papers also fall within the information-verifying paradigm as they automatically construct QA examples using extracted entities and relations and structured knowledge resources, like DrugBank.  Some examples include emrQA \cite{emrqa}, BioRead \cite{bioread}, BioMRC \cite{pappas-etal-2020-biomrc}, MedHop \cite{wikihop-medihop}. 
While these datasets enabled significant progress in machine comprehension, they include biases in questions that may not reflect real-world settings \cite{nq}.

\paragraph{Information-Seeking QA in General Domain}
Recognizing this challenge, others have followed an \textit{information-seeking} paradigm where the writer of questions is genuinely interested in finding the answer to the question, or at least does not have access to the answer. Examples of such datasets include WikiQA \cite{Yang2015WikiQAAC}, NewsQA \cite{newsqa}, MsMarco \cite{msmarco}, QuAC \cite{quac}, Natural Questions \cite{nq}, TyDiQA \cite{tydiqa}, and IIRC \cite{iirc}. Unlike \dataset, Natural Questions and TyDiQA\footnote{TyDiQA uses short snippets to prime annotators to write questions of interest, but the annotation process does not require workers to write questions grounded in those snippets.} questions are not grounded in any contexts, and the associated documents are linked to the questions after they are written.  In contrast, \dataset's questions are real follow-up questions about a paper that a reader of appropriate domain expertise would have after reading the title and the abstract. The priming lets the readers ask detailed questions that are specific to the papers in context, those that require a deeper understanding of the contexts, like those shown in Figure~\ref{fig:example} and Table~\ref{tab:qasper_examples_answer_evidence}. QuAC used similar data collection method but with focus on entities, which \dataset does not impose.

\paragraph{Domain-Specific Information-seeking QA}  Some work has been done on information-seeking QA on academic research papers. PubmedQA \cite{pubmedqa} derives Yes/No/Maybe questions from PubMed paper titles answered from the conclusion sections of the corresponding abstracts.  BioAsq benchmarks \cite{bioasq0, bioasq, bioasq1} focus on open-domain QA over PubMed abstracts. Like \dataset, BioAsq answers can take different forms (e.g., yes/no, extracted span(s)). \dataset differs from BioAsq in that questions are grounded in a single paper of interest.  Furthermore, \dataset uses the paper full text, not just the abstract. To the best of our knowledge, \dataset is the first information-seeking QA dataset in a computer science domain, while most prior work using academic research papers has been in biomedicine. Furthermore, with over 5K annotated questions, \dataset is also larger than other comparable human-annotated QA datasets -- PubmedQA and BioAsq contain 1K and 3.2K questions, respectively. Finally, \dataset poses a challenging full document-level task while other related datasets are abstract-level. Beyond the domain of academic research, realistic QA datasets have also been built in the privacy policy domain~\citep{ravichander-etal-2019-question,ahmad-etal-2020-policyqa}. These tasks are similar to our evidence selection task.

\section{Conclusion}
We presented \dataset, an information-seeking QA dataset over NLP research papers.  With natural questions asked as follow-up to titles and abstracts, the task presented by \dataset requires evidence from multiple paragraphs and/or figures and tables within the full text of the papers. Our empirical results show plenty of room for improvement when compared to the estimated human performance, and suggest that \dataset could serve as a test-bed for evaluating document-grounded QA research.

\section*{Ethical Considerations}
We present a new dataset that uses papers authored by other researchers.  To adhere to copyright, we have restricted ourselves to arXiv papers released under a CC-BY-* license, as identified via Unpaywall, which was used in the S2ORC \cite{lo-wang-2020-s2orc} dataset construction. Due to our choice to use arXiv as the source of papers, \dataset is almost entirely an English-language dataset, and QA systems built on \dataset would not be expected to work well on non-English language research papers.

We have determined the amount we paid the annotators to be well-above the minimum wage in our local area. While we do collect information about annotator background in NLP and familiarity with the papers they are annotating, we have not collected personal identifiable information without their permission except for payment purposes, and do not include any such information in the released dataset.

\bibliography{custom}
\bibliographystyle{acl_natbib}
\end{document}